\documentclass[conference]{IEEEtran}
\IEEEoverridecommandlockouts
\usepackage{cite}
\usepackage{amsmath,amssymb,amsfonts}
\usepackage{algorithmic}
\usepackage{graphicx}
\usepackage{textcomp}
\usepackage{xcolor}
\def\BibTeX{{\rm B\kern-.05em{\sc i\kern-.025em b}\kern-.08em
    T\kern-.1667em\lower.7ex\hbox{E}\kern-.125emX}}
    
\newcommand{\ignore}[1]{}
\usepackage{amsmath,amsfonts,bm}

\def\eqref#1{equation~\ref{#1}}
\def\1{\bm{1}}

\DeclareMathAlphabet{\mathsfit}{\encodingdefault}{\sfdefault}{m}{sl}
\SetMathAlphabet{\mathsfit}{bold}{\encodingdefault}{\sfdefault}{bx}{n}

\usepackage{microtype}
\usepackage{graphicx}
\usepackage{subfigure}

\usepackage{tabularx,booktabs}
\newcolumntype{C}{>{\centering\arraybackslash}X} % centered version of "X" type
\usepackage{lipsum}

\def\w{{\bf w}}

\def\b{{\bf b}}

\def\x{{\bf x}}

\begin{document}

\title{Pairwise Margin Maximization \\ for Deep Neural Networks\\
%\title{Pairwise Margin Maximization Regularization Term for Deep Network\\
}

%\author{\IEEEauthorblockN{1\textsuperscript{st} Berry Weinstein}
\author{\IEEEauthorblockN{Berry Weinstein}
\IEEEauthorblockA{\textit{School of Computer Science}\\
Reichman University\\ 
%Herzliya, Isreal \\
berry.weinstein@post.idc.ac.il}
\and
%\IEEEauthorblockN{2\textsuperscript{nd} Shai Fine}
\IEEEauthorblockN{Shai Fine}
\IEEEauthorblockA{\textit{Data Science Institute}\\
Reichman University \\
%Herzliya, Isreal \\
shai.fine@idc.ac.il}
\and
%\IEEEauthorblockN{2\textsuperscript{rd} Yacov Hel-Or}
\IEEEauthorblockN{Yacov Hel-Or}
\IEEEauthorblockA{\textit{School of Computer Science}\\
Reichman University \\
%Herzliya, Isreal \\
toky@idc.ac.il}
}

\maketitle

\begin{abstract}
The weight decay regularization term is widely used during training to constrain expressivity, avoid overfitting, and improve generalization. Historically, this concept was borrowed from the SVM maximum margin principle and extended to multi-class deep networks. Carefully inspecting this principle reveals that
it is not optimal for multi-class classification in general, and in particular when using deep neural networks. In this paper, we explain why this commonly used principle is not optimal and propose a new regularization scheme, called  {\em Pairwise Margin Maximization} (PMM), which measures the minimal amount of displacement an instance should take until its predicted classification is switched. In deep neural networks, PMM can be implemented in the vector space before the network's output layer, i.e., in the deep feature space, where we add an additional normalization term to avoid convergence to a trivial solution. We demonstrate empirically a substantial improvement when training a deep neural network with PMM compared to the standard regularization terms.
\end{abstract}

\section{Introduction}

Over the last decade, deep neural networks (DNNs) have become the {\it machine learning} method of choice in a variety of applications,  demonstrating outstanding performance, often close to or even better than human-level performance.
Nevertheless, some researchers have shown that DNNs  can generalize poorly even with small data transformations \cite{azulay2018deep} as well as overfit arbitrarily corrupted data \cite{zhang2017multi}. Additionally, problems such as adversarial attacks, which cause neural networks to misclassify slightly perturbed input data, can be a source of concern in real-world scenarios (cf. \cite{szegedy2013intriguing, goodfellow2014explaining}). 
These challenges have motivated researchers to investigate whether properties that enabled classical machine learning algorithms to overcome the above-mentioned problems can be useful in helping DNNs resolve similar issues. 
Specifically, for linear classifiers, it has been evident that a classifier with a large margin over different classes in the training data produced better generalization results as well as stronger robustness to input perturbations \cite{bousquet2001algorithmic}. 
%The large margin principle, 
The maximal margin principle, i.e., maximizing the smallest distance from the instances to the classification boundary in the feature space, has played an important role in theoretical analysis of generalization, and helped to achieve remarkable practical results \cite{cortes1995support},  
%It has been evident that a classifier with a large margin over different classes in the training data produces better generalization results 
as well as robustness to input perturbations \cite{bousquet2001algorithmic}. 
Of particular interest to our study are the extensions to multi-class classification: multi-class perceptron (see Kesler’s construction,~\cite{duda1973pattern}); multi-class SVM \cite{vapnik1998statistical}; multi-class margin distribution \cite{zhang2017multi}, and the mistake-bound for multi-class linear separability that scales with $(R/\gamma)^2$, where $R$ is the maximal norm of the samples in the feature space, and $\gamma$ is the margin~\cite{crammer2003ultraconservative}. %

\ignore{
Recently, \cite{jiang2019predicting} developed a measure for predicting the generalization gap~\footnote{The difference in accuracy between training and testing performance.} in DNNs that leverages {\it margin distribution}~\footnote{The distribution of distances to the {\it decision boundaries}.} \cite{garg2002generalization} as a more robust assessment for the margin notion in DNNs. In their work, they also point out that this measure can be used as an auxiliary loss function to achieve better generalization.
}

In all the above studies, the multi-class extension of the %large margin 
maximal margin principle is realized by the weight decay regularization scheme, which is a natural extension of the regularization term in the binary SVM case.
We argue here that the standard weight decay regularization is not optimal in the multi-class case. In particular, we show that the weight decay term aims at maximizing the margins along the {\em one-vs-all} decision boundaries. According to the way a multi-class classification prediction is implemented, however, these decision boundaries are less of interest and the focus should be on using the {\em one-vs-one} decision boundaries as the margins. 
For the one-vs-one decision boundaries, we present a novel regularization term, which we call {\it Pairwise Margin Maximization} (PMM). The PMM regularization term maximizes the margins around the one-vs-one boundaries and can be added to any loss. 
%empirical risk loss. 
We derive the regularization term starting from the first principle in the binary case and generalize it to the multi-class setting.

%Furthermore, since we apply our method to the feature space and since it is dynamic in the course of training of neural networks, we address this by scaling our formulation by the radius of the feature space, i.e., the maximal norm of the samples in the feature space. 
In systems where multi-class classification uses DNNs, we propose an extension that applies the maximal margin principle in the feature space rather than in the input space. Nevertheless, since the feature space is learnable, the margins can be trivially maximized by scaling the feature space. We address this issue by scaling our formulation by the radius of the feature space, i.e., the maximal norm of the samples in the feature space. This normalization is similar in spirit to the normalization proposed in~\cite{crammer2003ultraconservative}.
We applied PMM on the last layer of various DNNs, in different classification tasks and in various domains. We empirically show that the PMM scheme provides a substantial improvement in accuracy while maximizing the margins between pairwise classes. In particular, we achieve considerable accuracy improvement in various image and text classification tasks, including CIFAR10, CIFAR100, ImageNet, MNLI, and QQP.

\ignore{
To conclude, our contribution is that we revisit the weight decay regularization scheme as a driving force for margin maximization. We show that the commonly used margin maximization is applied to the wrong decision boundaries and thus, it produces non-optimal generalization. We propose a different set of decision boundaries (one-vs-one) to enforce upon them the margin maximization and introduce a new regularization term that optimizes this concept. We show both visually and empirically that this scheme improves the separation between classes as well as the accuracy over a large set of classification problems.
}

To conclude, our contribution is a novel regularization term, specifically designed to enforce the maximization of the margin on the one-vs-one decision boundaries, rather than maximizing the margin on the one-vs-all decision boundaries, which is derived by the weight decay regularization scheme.
We show both visually and empirically that this scheme improves the separation between classes as well as the accuracy over a large set of classification problems.

\ignore{
Our contribution is twofold. Alongside improving generalization performance using a new regularization scheme, we leverage the large margin principle to improve convergence during training using a selective-sampling scheme and assisted by a measure that we call the {\it Minimal Margin Score} (MMS). 
Essentially, MMS measures the distance to the decision boundary of the two most competitive predicted labels. 
%Specifically, we suggest a method 
This measure, in turn, is used 
to select, at the back-propagation pass, only those instances that accelerate %the training
convergence, thus speeding up the entire training process. 
It is worth noting that
our selection criterion is based on computations that are 
%calculated in any event as  
an integral part of the forward pass, thus taking advantage of the "cheaper" inference computations. Lastly, we empirically show that using the MMS selection scheme with a faster learning-rate regime can improve, to a large extent, the convergence process. 
}

\subsection{Previous Approaches}
%The large margin 
The maximal margin principle has proven to be fundamentally important in machine learning as it was shown to correlate with generalization and accuracy \cite{schapire1998boosting}. Although most efforts revolved around binary classification, extensions to multi-class classification were also suggested \cite{vapnik1998statistical, crammer2001, zhang2017multi,crammer2003ultraconservative}.
Soudry et al. \cite{soudry2018implicit} proved that cross-entropy loss in linear DNNs, together with {\it stochastic gradient descent} (SGD) optimization, converges to the maximal margin solution. This proof, however, was not extended to nonlinear DNNs, and indeed, Sun et al. \cite{Sun2015LargeMD} affirmed that cross-entropy alone is not enough to achieve the maximal margin in non-linear DNNs and that an additional regularization term is needed.  

Margins in the input space using nonlinear DNNs can be approximated using derivatives assisted by the back-propagation scheme. Elsayed et al. \cite{elsayed2018large} presented a multi-class linear approximation of the margins as an alternative loss function. They applied their margin-based loss at each and every layer of the neural network. Moreover, their method requires a second-order derivative computation due to the presence of first order gradients in the loss function itself. Explicit computation of the second order gradients for each layer of the neural network, however, can be quite expensive, especially when DNNs are getting wider and deeper. To address this limitation, they used a first-order linear approximation to deploy their loss function more effectively. 
Later,  Jiang et al. \cite{jiang2019predicting} presented a 
%measure, based on the SVM margin 
margin-based measure
that strongly correlates with the generalization gap in DNNs. Essentially, they measured the difference between the training and the test %capabilities of a neural network by obtaining statistical information based on 
performances of a neural network using marginal distribution statistics \cite{garg2002generalization}. Sokoli{\'c} et al. \cite{sokolic2017robust} used the input layer to approximate the margin via the network's Jacobian matrix and showed that maximizing their approximations leads to a better generalization. In contrast, we show that applying our margin-based regularization to the output layer alone achieves substantial improvement. 

The rest of the paper is organized as follows. In Section \ref{sec:MMS} we explain the need for PMM in deep classification problems with the classical maximal margin principle in binary and multi-class settings. In Section \ref{MMP_details} we derive the PMM regularization scheme for DNNs and show preliminary results on a simple classification task using the CIFAR10 dataset. We expand the experimental results in Section \ref{sec:Exp} by applying PMM to additional vision tasks on various CNN architectures as well as binary {\it natural language processing} (NLP) classification tasks on the BERT\textsubscript{BASE} model. Lastly, we summarize our contributions in Section \ref{sec:Discussion}.

\section{Margin Analysis for Binary and multi-class Classification}
\label{sec:MMS}

The maximal margin principle is traditionally presented in the context of a shallow linear classifier \cite{vapnik1998statistical, crammer2001}. %The large margin 
It is the core principle behind the {\it support vector machine} (SVM) classifier. It was shown that maximizing the margin between the data samples and the decision boundary also maximizes the classifier's generalization capabilities \cite{vapnik1998statistical}. In the original work by Vapnik \cite{cortes1995support}, the maximal margin principle was applied to the data points closest to the boundary (support vectors), while later works \cite{zhang2019optimal} extended the maximal margin principle to the mean and variance of the distances. 

We start our discussion with the classical derivation of the maximal margin principle and its traditional extension to multi-class case. 
We then show that this extension is not optimal as it refers to the  one-vs-all maximal margin principle. Next, we suggest a new maximal margin principle that we term {\em Pairwise  Margin Maximization} (PMM). In fact, the PMM principle emerged from the maximal margin principle where the one-vs-one classification scheme is  carried out. Lastly, we describe the necessary components for adapting PMM to DNNs.  

Consider a classification problem with two classes ${\cal Y} \in \{ +1,-1 \}$. 
We denote by ${\cal X} \in {\cal R}^d$ the input space. 
Let $f(\w^T \x + b)$ be a linear classifier, where $\x \in {\cal X}$ and 
$$
f(z)= \left\{
\begin{array}{ll}
 +1 &~~\mbox{if}~~ z \geq 0 \\
 -1	&~~\mbox{otherwise} 
 \end{array}
 \right.
$$

The classifier is trained using a set of examples $\{(\x_1,y_1),(\x_2,y_2),\cdots, (\x_n,y_n)\} \in ({\cal X} \times {\cal Y})^n$ 
where each example is sampled identically and independently from an unknown distribution ${\cal D}$ over ${\cal X}$. 
The goal is to classify correctly new samples drawn from ${\cal D}$.
%, i.e. $y f(\w^T \x) \geq 0,~~(\x,y)\in{\cal D}$.
% In order to train the classifier, we require that each example from the training set will be classified correctly, namely: $y_i f(\w^T \x_i) \geq 0$. However, its capability to correctly classify new data, drawn from ${\cal D}$, is shown to be correlated with minimal margin of the training set to the decision boundary.

Denote by $\ell$ the (linear) decision boundary defined by the classifier $f$:
\begin{equation}
\label{eq:ell}
\ell = \{ \x ~|~ \w^T \x + b = 0\} 
\end{equation}
The geometric distance of a point $\x$ from $\ell$ is given by 
\begin{equation}
\label{eq:d_i}
d(\x) = \frac{\w^T \x +b}{\| \w \|}
\end{equation}
For a linearly separable training set,
the maximum margin classifier demonstrates the best generalization capability, which is achieved by selecting the classifier that maximizes the margin $\hat d$ \cite{cortes1995support}:
%However, in order to increase the capacity (generalization capability) of $f$, 
%the SVM with hard margin aims at maximizing the minimum margin $\hat d$
$$
\hat d = \arg \max_{\w,b}  d ~~~\mbox{s.t.}~~~y_i \frac{\w^T \x_i +b}{\| \w \|} \geq  d,  ~~~~\forall i=1,\cdots,n
$$
This optimization is redundant with the length of $\w$ and $b$. 
Namely, if $(\w^*,b^*)$ is the optimal solution, then so is $(\alpha \w^*, \alpha b^*)$.
%In order to remove the redundancy,  we impose $y_i (\w^T \x_i +b) \geq 1$ which is translated into the following equivalent minimization problem~\cite{cortes1995support}
Imposing $y_i (\w^T \x_i +b) \geq 1$ removes this redundancy and results in the following equivalent minimization problem~\cite{cortes1995support}:
$$
\min_{\w,b} \|\w\|^2 ~~~\mbox{s.t.}~~~ y_i (\w^T \x_i +b) \geq 1,
~~~~\forall i=1,\cdots,n $$
The above optimization forces all samples to be classified correctly (with a margin no less than 1). 
%For noisy data, or a data set that is not linearly separable, the set of linear constraints are relaxed and substituted with the hinge loss, i.e. soft margin 
To handle noisy and linearly inseparable data, the set of linear constraints is relaxed using soft margins by replacing it with the hinge loss,
\begin{equation}
\label{eq:min}
    \min_{\w,b} \|\w\|^2+ \lambda \sum_i \max (0, 1- y_i (\w^T \x_i +b))
\end{equation}
The two terms in the above minimization problem employ two complementary forces. The left term is the {\em regularization} component and it promotes increasing the margin between the data points and the decision boundary, thus improving the generalization capability.  The right term of the formula is the {\it empirical risk} component, promoting correct classification of the training samples. 

We now extend the %large margin principle 
maximal margin principle to the multi-class case. Let us assume that we have a classification problem with $k$ classes, ${\cal Y} \in \{1,\cdots,k \}$, and a set of $n$ training samples: $\{ (\x_i, y_i)\} \in ({\cal X} \times {\cal Y})^n$. 
For a given input $\x$, a trained classifier assigns a set of scores, i.e., a score to each class: $s_j(\x): {\cal X}\rightarrow \mathbb{R}, ~\forall~j \in {\cal Y}$. For a linear classification, the $j^{th}$ score of instance $\x$ is:
$$
s_j(\x) = \w_j^T \x + b_j
$$
The predicted class is then chosen by the maximal score attained over all classes,
$$
{\hat y} = \arg \max_{j \in {\cal Y}} s_j(\x)
$$
For a training pair $(\x_i, y_i)$, denote by $s_{y_i}(\x_i)$ the score attained for the true class of $\x_i$ and by $s_{m_i}(\x_i)$ the maximal score attained for the non-true classes, i.e., $m_i$ is the %{\it competitive} class vis-\`{a}-vis $y_i$:
most {\it competitive} class with respect to $y_i$:
$${m_i} = \arg \max_{j \neq y_i} s_j(\x_i)$$
We define:
$$
\xi_i= s_{y_i}(\x_i) - s_{m_i}(\x_i) 
$$
 $\xi_i$ is the difference between the true score and the score of the {\it competitive} class. When $\xi_i$ is positive, the true class attains the best score; otherwise $\xi_i$ is negative. The larger $\xi_i$ is, the larger the margin we have between the true and the competitive scores.

The commonly used multi-class classification scheme is defined as a one-vs-all classification scheme, where the maximal margin principle of Equation~\ref{eq:min} is generalized to the multi-class \cite{crammer2001}:
\begin{equation}
\label{eq:min_mc2}
\min_{W,b} \sum_{j=1}^k \| \w_j \|^2 + \lambda \sum_{i=1}^n \max(0, 1-\xi_i) 
\end{equation}
%
%\begin{eqnarray}
%\label{eq:min_mc1}
%&\min_{(W,b,\xi)} \sum_{j=1}^k \| \w_j \|^2 + \lambda \sum_{i=1}^n \xi_i \\
%&\mbox{s.t.}~ \forall i \in [n], \forall j \in [k], j \neq y_i: ~s_{y_i}(\x_i)- s_j(\x_i) \geq 1-\xi_i \nonumber
%\end{eqnarray}
In the above minimization, the optimization is over $W,b \doteq \{(\w_j,b_j)\}_{j=1}^k$.
%Denote by $s_{m_i}(\x_i)$ the maximal score attained for the non-true classes, 
%i.e., $m_i = \arg \max_{j \neq y_i} s_j(\x_i)$, and define:
%$$
%\eta_i= s_{y_i}(\x_i) - s_{m_i}(\x_i) 
%$$
%i.e., $\eta_i$ is the minimal difference between $s_{y_i}(\x_i)$ and any other score, where class $m_i$ is the {\it competitive} class vis-\`{a}-vis $y_i$. 
%Using the above definitions, the minimization of Equation~\ref{eq:min_mc1} can be simplified:
%\begin{equation}
%\label{eq:min_mc2}
%\min_{W,b} \sum_{j=1}^k \| \w_j \|^2 + \lambda \sum_{i=1}^n \max(0, 1-\eta_i) 
%\end{equation}
The left term, as in the binary case, refers to the regularization term and emerges from the maximum margin principle. Since it minimizes $\| \w_j\|^2 $, $j=1..k$, it promotes large margins along the $k$ decision boundaries:
$$
\ell_j = \{ \x ~|~ \w_j^T \x + b_j=0\},~~~~ j=1..k
$$
Nevertheless, the desired property is not the way in which the predicted class is evaluated, maximizing the margin along each $\ell_j$.
Rather, our hypothesis is that the values of $\xi_i$ should be maximized, i.e., the difference between the distance $\x_i$ to $\ell_{y_i}$ and its distance to $\ell_{m_i}$. In other words, the regularization term, as defined in Equation~\ref{eq:min_mc2}, does not maximize the correct margins. 

\begin{figure}[tbh]
% \vspace{-1em}
    \centering
    \includegraphics[width=0.4\textwidth]{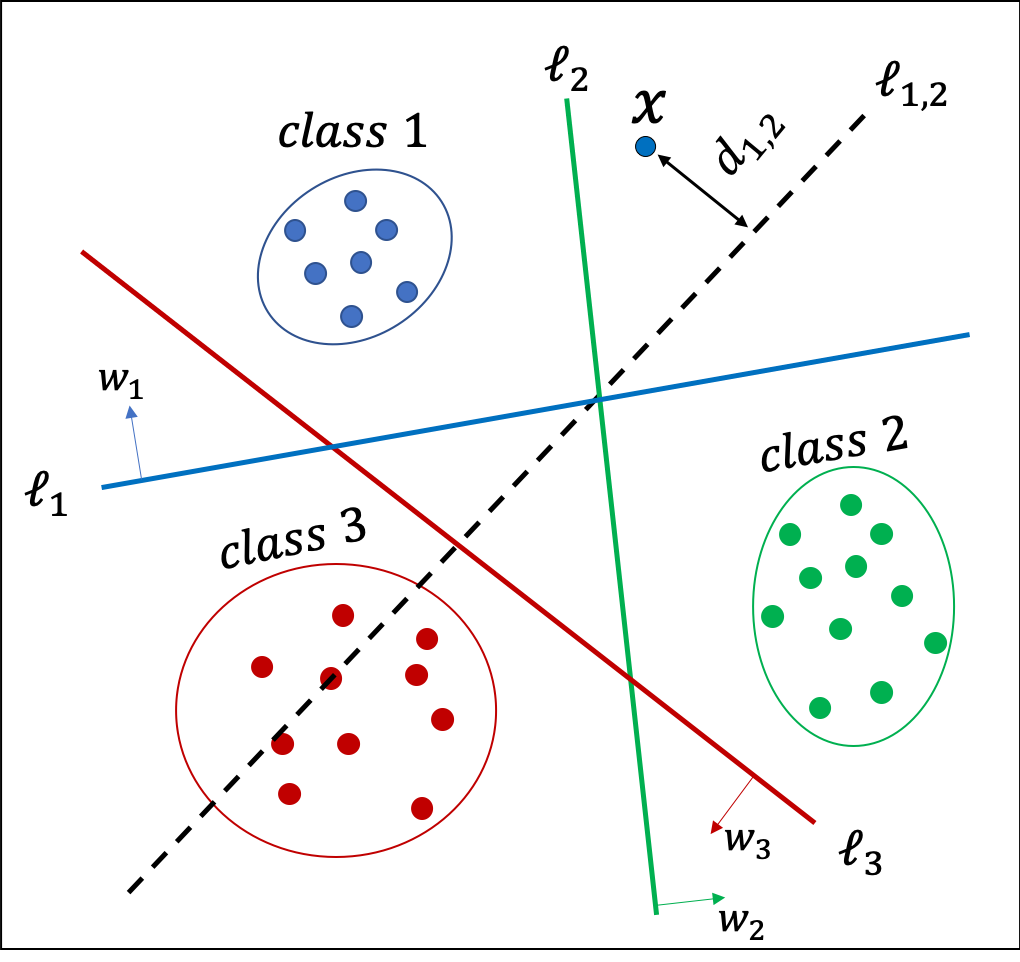}
    \caption{Illustrative example of pairwise decision boundaries. Three classes along with their one-vs-all decision boundaries are presented.
    The true class of sample point $\x$ is class 1 (the blue class) while the most competitive class for this point is class 2 (the green class). Thus, the margin in question is in respect to $\ell_{1,2}$ whose distance to $\x$ is $d_{1,2}$.
}
    \label{fig:mms_new}
\end{figure}

To clarify this point, consider the illustrative example in Figure~\ref{fig:mms_new}. A data point denoted by $\x$ is given as a training input. Assume the true class of this data point is class 1 (the blue class). The most competitive class to class 1 is class 2 (the green class). Consider the margin in the space: 
$\ell_{1,2}= \{\x ~|~ s_1(\x)=s_2(\x) \} $.
The line $\ell_{1,2}$ splits the space into two half spaces where one side includes the data points closer to $\ell_1$ and, in the other side, the points are closer to $\ell_2$. In Figure~\ref{fig:mms_new}, the point $\x$ is located on the right side of $\ell_{1,2}$ and will be classified correctly since $s_1(\x)>s_2(\x)$. If, however, we maximize the margins around all of $\ell_i$, $i=1..3$ (the one-vs-all boundaries), the point $\x$ is located on the wrong side of $\ell_2$ and the parameters $\w_2,b_2$ will be updated to maximize the margin of  $\ell_2$, although the predicted classification of $\x$ is correct and the margin width is satisfied.

In the following we suggest an alternative approach that is derived and justified directly from maximization of $\xi_i$. We start with the observation that although we deal with $k$ decision boundaries, these boundaries can induce one-vs-one boundaries as well. For any two classes, $(p,q) \in {\cal Y} \times {\cal Y}$, the pairwise decision boundary between $p$ and $q$ is given by (see Figure~\ref{fig:mms_new}):
$$
\ell_{p,q} = \{\x ~|~ s_p(\x)=s_q(\x) \} =
\{ \x ~|~  \w_p^T \x + b_p = \w_q^T \x + b_q \}.
$$
Denoting $\w_{p,q}=\w_p -\w_q$ and $b_{p,q}=b_p -b_q$, the decision boundary $\ell_{p,q}$ can be rewritten as:
$$
\ell_{p,q} = \{\x ~|~ \w_{p,q} \x + b_{p,q} =0 \} 
$$
which is similar to the binary case in Equation~\ref{eq:ell} where $\w_{p,q}$ replaces $\w$ and $b_{p,q}$ replaces $b$. In fact, although we optimize only for $k$ classification parameters ($(\w_i, b_i)$, $i=1..k$), the resulting parameters can be interpreted as $k(k-1)/2$ decision boundaries that are associated with all pairwise one-vs-one classifications. 

Similarly to Equation~\ref{eq:d_i}, the geometric distance of a point $\x$ from $\ell_{p,q}$ is (see Figure~\ref{fig:mms_new}):
\begin{equation}
\label{eq:dpq}
d_{p,q}(\x) =  \frac{\w_{p,q}^T \x +b_{p,q}}{\| \w_{p,q} \|}
\end{equation}
For point $\x_i$, the decision boundary between $y_i$ and its most competitive class $m_i$ is $\ell_{y_i,m_i}$, whose geometric distance to $\x_i$ is
\begin{equation}
\label{eq:d2}
d_{y_i,m_i}(\x_i) =  \frac{\w_{y_i,m_i}^T \x_i +b_{y_i,m_i}}{\| \w_{y_i,m_i} \|} 
\end{equation}
%where $\w_{y_i,m_i}=\w_{y_i}-\w_{m_i}$ and $b_{y_i,m_i}=b_{y_i}-b_{m_i}$.
Thus, our goal is to maximize the margins around $\ell_{y_i,m_i}$, for $i=1..n$.  Note that $d_{y_i,m_i}(\x_i)$ is non-negative if the classification is correct ($s_{y_i}(\x_i) \geq s_{m_i}(\x_i)$) and negative otherwise.

\ignore{
An illustrative example for this margin is given in Figure~\ref{fig:mms_new}. Assume the true class of the sample point denoted by $\x$ is class 1 (the blue class). The most competitive class to class 1 is class 2 (the green class). Therefore, the margin in question is about $\ell_{1,2}$. And indeed, in this example, the point $\x$ is located at the right side about $\ell_{1,2}$ and it will be classified correctly. Moreover, if $d_{1,2}\geq 1$ the loss for this points will be 0. On the the other hand, if we maximize the margins about $\ell_i$, the point $\x$ is located at the wrong side of $\ell_2$ and the parameters $\w_2,b_2$ will be updated to maximize the margin of  $\ell_2$, although the predicted classification of $\x$ is correct and the margin width is satisfied.
}
%We continue the derivations similarly to the binary case. For the multi-class case, Equation~\ref{eq:min} can be generalized into the following optimization problem:
Following the above justification, the same procedure applied to Equation~\ref{eq:d_i} to derive Equation~\ref{eq:min} in the binary case can be applied here as well. Thus, Equation~\ref{eq:d2} yields: 
\begin{equation}
\label{eq:optim2}
    \min_{W,b}  \sum_i \|\w_{y_i,m_i}\|^2+ \lambda \sum_i \max (0, 1- \xi_i)
\end{equation}
where, 
$$\xi_i=\w_{y_i,m_i}^T \x_i +b_{y_i,m_i} = s_{y_i}(\x_i) - s_{m_i}(\x_i)$$ 
as defined above. 

The above regularization term aims at increasing $d_{y_i,m_i}(\x_i)$, namely, the margins along $\ell_{y_i,m_i}$, as desired. In contrast to Equation~\ref{eq:min_mc2} and the common $L_2$ regularization scheme, where the Frobenius norm of $W$ is minimized: i.e., $\sum_{j=1}^k \|\w_j\|^2$, Equation~\ref{eq:optim2} minimizes the pairwise margins $\sum_i \|\w_{y_i,m_i}\|^2$, { which is a different regularization objective}. For a pair of classes $(i,j)$, the standard $L_2$ regularization minimizes $\|\w_i\|^2+\|\w_j\|^2$ while the suggested scheme minimizes $\|\w_i\|^2+\|\w_j\|^2 - \w_i^T \w_j$. Since the margins are defined over each pairwise boundary, we call this regularization scheme a {\em pairwise margin maximization} (PMM).

Another point to note here is that the summation in Equation~\ref{eq:optim2} is performed over the instance points ($i$ is the instance index). This means that the larger the existence of the pair $(i,j)$ as a competitive pair, the stronger the applied regularization. This can be interpreted as a minimization over the {\em margin distribution} rather than the maximal margin per se. If the instances are evenly distributed over the classes, then this is equivalent to summation over the class pairs. Otherwise, this summation compensates for class imbalance in the regularization term. Additionally, the PMM regularization term is applied without any additional computational cost since the pairwise terms are computed per example and the entire $O(k^2)$ pairwise terms are not necessarily computed.

\section{Maximal Margin in Deep Networks}
\label{MMP_details}
Applying PMM directly to DNNs poses several problems. First, these networks employ a nonlinear mapping from the input space into a feature space: 
%$\phi_i= m(\x_i): {\cal X} \rightarrow {\Phi}$.  
$\phi_i= F(\x_i, \theta): {\cal X} \rightarrow {\Phi}$, where $\theta$ are the network's parameters.  
The  vector $\phi_i$  can be interpreted as a feature vector based on which the last layer in the network calculates the scores, for each class, via a fully-connected layer,  $s_j(\phi_i)=\w_j^T \phi_i + b_j$. 
Maximizing the margin in the input space ${\cal X}$, as suggested in \cite{sokolic2017robust}, requires back-propagating derivatives along the network up to the input layer, and calculating distances to the boundary up to the first order approximation. In highly nonlinear mappings, this approximation becomes inaccurate very fast as we move away from the decision boundary. 

To overcome this problem, our scheme maximizes the margin only in the last layer, where the distances to the decision boundary are Euclidean in the {\em feature space} $\Phi$:
\begin{equation}
\label{eq:d3}
    d_{y_i,m_i}(\phi_i) =  \frac{\w_{y_i,m_i}^T \phi_i +b_{y_i,m_i}}{\| \w_{y_i,m_i} \|} 
\end{equation}

Working with the feature space, however, presents a new challenge because the feature space $\Phi$ can be modified in the course of training. 
%This is different than the input space ${\cal X}$ (Equation~\ref{eq:d2})  which is fixed. 
This is different than maximizing the margins in the input space ${\cal X}$ (c.f. Equation~\ref{eq:d2}) or in a kernel-induced feature space (e.g., SVM), since in both cases the space is fixed. 
%Accordingly, 
If the feature space keeps changing, then maximizing the margins in Equation \ref{eq:d3} can be trivially attained by scaling up the feature space $\Phi$. 

To avoid the trivial solution, we must normalize the feature space $\Phi$. In our scheme, we divide Equation \ref{eq:d3} by $\| \phi_{max}\|$, which is 
the maximal norm of the samples (in the feature space) in the current batch.
%which is the sample with the maximal feature magnitude at the current batch. 
This ensures that scaling up the feature space will not increase the distance arbitrarily. 
%The proposed formulation is translated, similarly to Equation~\ref{eq:optim2}, into the following optimization problem 
Putting all the components of our scheme together, we end up with the following optimization problem: 
\begin{equation}
\label{eq:d4}
\min_{W,\b}  \sum_i {\cal R}_i + \lambda \sum_i {\cal C}_i
\end{equation}
where 
$${\cal R}_i = \|\w_{y_i,m_i}\|^2 \|\phi_{max}\|^2 $$
is the pairwise regularization term, and ${\cal C}_i$ is the empirical risk term: 
$$
{\cal C}_i = \max(0, 1 -\xi_i)
$$
% \begin{equation}
% \label{eq:hinge}
% {\cal C}_i = \max (0, 1- (\w_{y_i,m_i}^T \phi_i + b_{y_i,m_i}))
% \end{equation}
% is the empirical risk term composed of hinge loss.
% \begin{equation}
% \label{eq:optim3}
%     \min_{W,\b}  \sum_i \|\w_{y_i,m_i}\|^2 \|\phi_i\|^2 + \lambda \sum_i \max (0, 1- (\w_{y_i,m_i}^T \x_i +b_{y_i,m_i}))
% \end{equation}

Finally, for DNNs, better classification results are commonly obtained using cross-entropy rather than hinge loss. Our formulation supports employing cross-entropy as well. The empirical risk term is simply replaced with 
\begin{equation}
{\cal C}_i = - \log (P_{y_i})
\end{equation}
 where $P_{y_i}$ is the conditional probability of the true label $y_i$ as obtained from the network after the softmax layer:
 $$
 P_{y_i}=\frac{e^{s_{y_i}(\x_i)}}{\sum_j e^{s_j(\x_i)} }
 $$
 Similarly to the hinge loss formulation, the cross-entropy term will strive for correct classification while the regularization term will maximize the margin. 
\ignore{
 The last issue while implementing the above formulation in deep networks is that hinge loss provides inferior results compared to  the cross-entropy. 
 \ignore{
 This was demonstrated over a large set of problems \cite{XX} and in fact, the common practice today is to apply only cross-entropy criterion in DNN solving multi-class problems.  However, this issue can be easily mitigated by replacing the hinge loss in Equation~\ref{eq:optim2} with  cross entropy: 
\begin{equation}
{\cal C}_i = - \log (P_{y_i})
\end{equation}
 where $P_{y_i}$ is the conditional probability of the true label $y_i$ as obtained from the network after the softmax layer.  Similarly to  hinge loss, the cross entropy will promote the correct classification while the regularization term will maximize the margin. 
 }
 While for SVM, hinge loss is commonly used as the empirical risk, in deep networks the common practice is to use cross-entropy
 %, i.e. ${\cal C}_i = - \log (P_{y_i})$, 
$$
{\cal C}_i = - \log (P_{y_i})
$$
 where $P_{y_i}$ is the predicted probability of the true label $y_i$ obtained from the network after the softmax layer: 
 $$
 P_{y_i}=\frac{e^{s_{y_i}(\x_i)}}{\sum_j e^{s_j(\x_i)} }
 $$
 Similarly to hinge loss, cross-entropy will strive for correct classification while the regularization term will maximize the margin. 
 }
 
 Note that the regularization term in this scheme is different from the weight decay commonly applied in DNNs. First, the minimization is applied over the differences: $\|\w_{y_i,m_i}\|^2=\|\w_{y_i} -\w_{m_i}\|^2$. Next, the regularization term is multiplied by the $\|\phi_{max}\|$.
 Lastly, the regularization term is implemented only at the last layer. 
 %and it does not applied to each layer independently.  

\begin{figure}[tbh]
% \vspace{-1em}
    \centering
         \includegraphics[width=1\columnwidth]{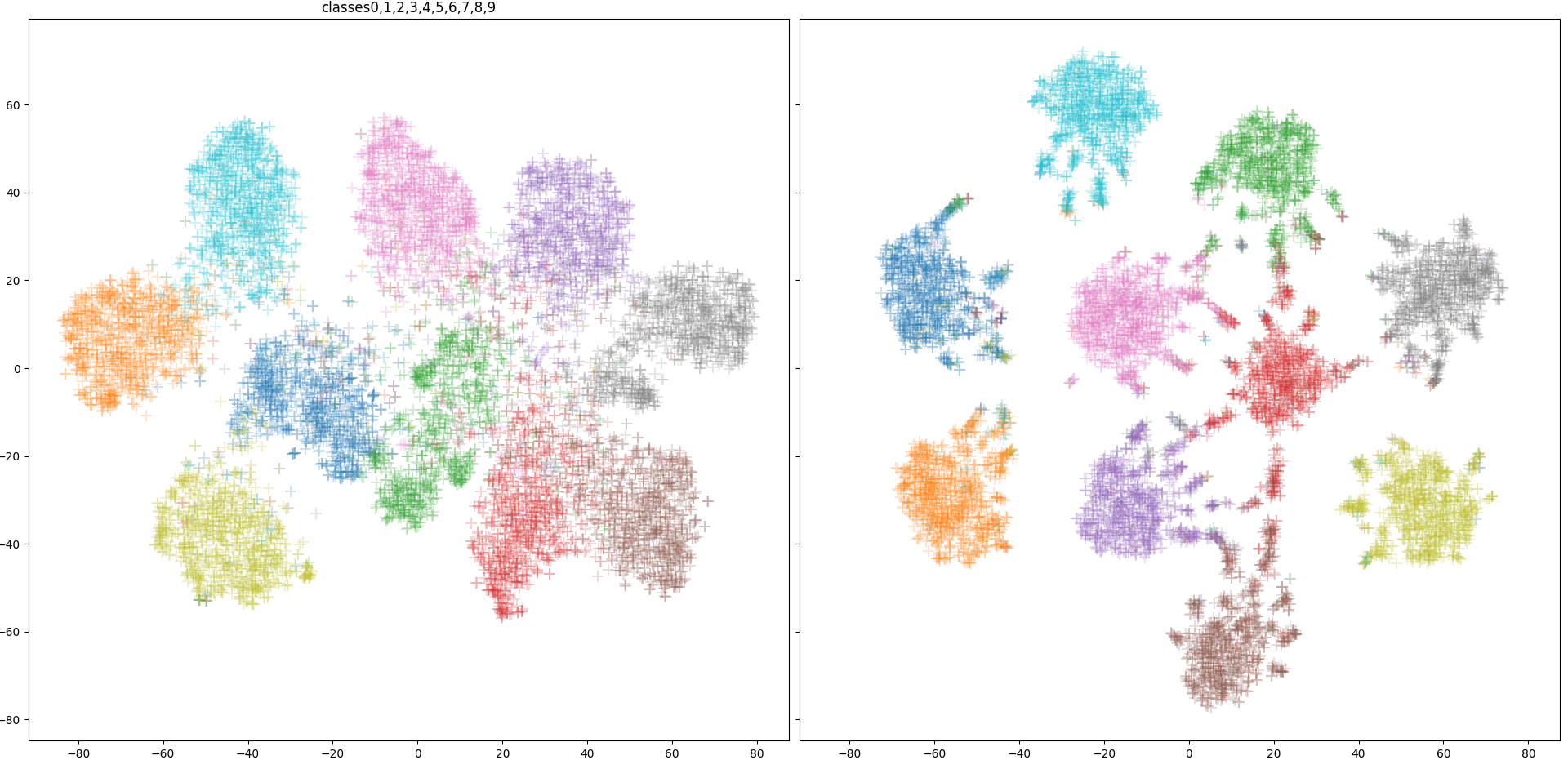}  
    % \qqqqqquad %add desired spacing between images, e. g. ~, \quad, \qquad, \hfill etc. 
        
     \caption{A scatter plot of the last layer in ResNet44 trained on CIFAR-10 after embedding into 2D space using t-SNE. Left: Baseline using $L_2$ regularization. Right: PMM using the suggested scheme.}
    \label{TSNE_cifar10_new}
    % \vspace{-1em}
\end{figure}

The effect of the PMM regularization scheme compared to the baseline $L_2$ regularization is demonstrated in Figure~\ref{TSNE_cifar10_new}. In this plot we consider the feature points in the penultimate layer of the ResNet44 (before the fully connected layer) trained on the CIFAR-10 data set. The features were %dimensionality reduced 
projected to a lower dimension
using t-SNE \cite{van2008visualizing} and visualized in a 2D scatter plot. Each class is indicated by a different color. The left plot presents the feature distribution using the standard $L_2$ regularization while the right plot shows the distribution using the PMM scheme. It is demonstrated that the PMM plot presents well-separated clusters with large margins between each class while the baseline plot indicates overlapping classes with small margins. 

\section{Experiments}
\label{sec:Exp}

\ignore{
\begin{figure*}[tbh]
% \vspace{-1em}
    \centering
         \includegraphics[width=2\columnwidth]{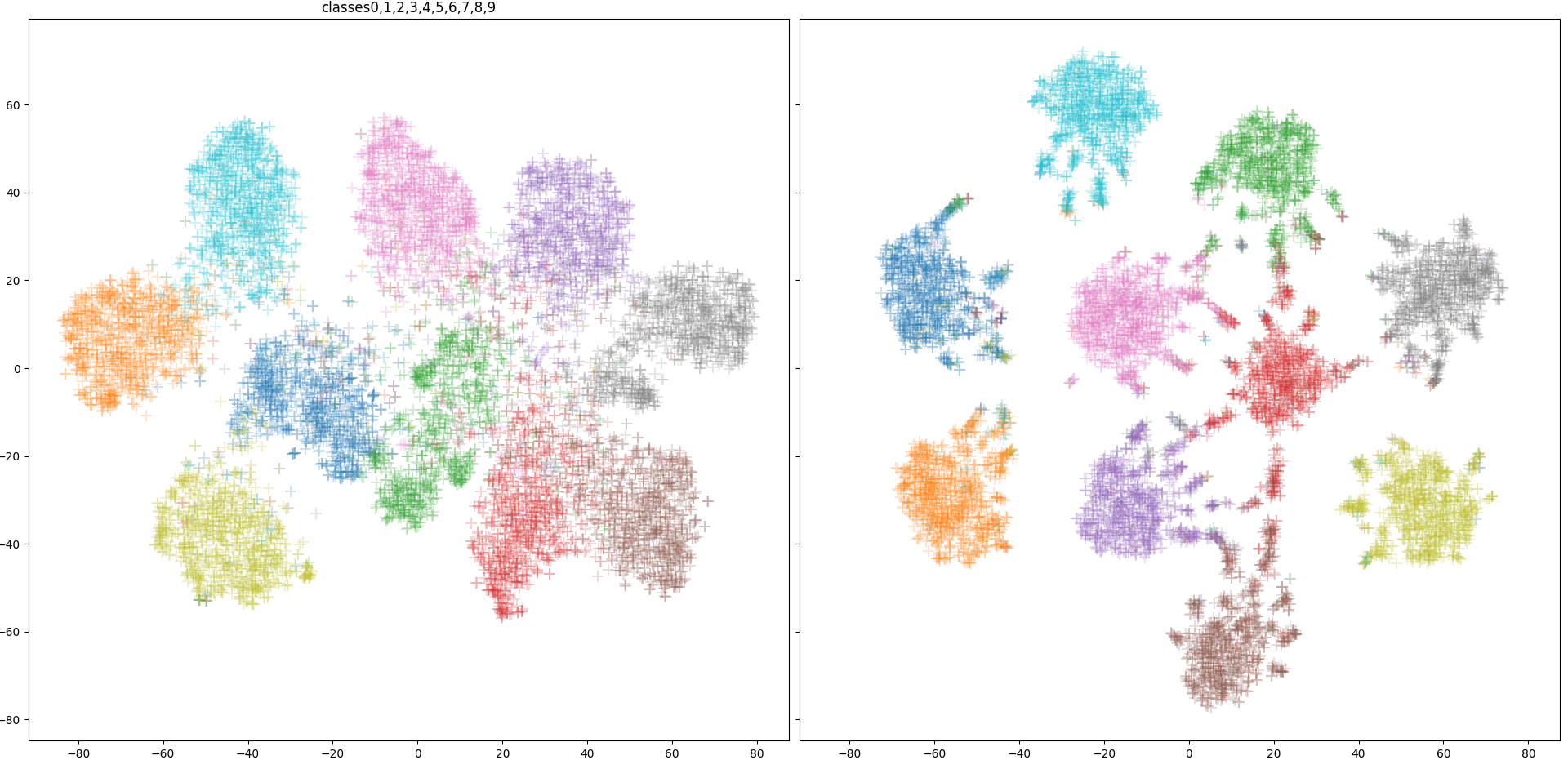}  
    % \qqqqqquad %add desired spacing between images, e. g. ~, \quad, \qquad, \hfill etc. 
        
     \caption{TSNE on ResNet44 and Cifar10. Baseline (upper) vs PMM (lower).}
    \label{TSNE_cifar10}
    % \vspace{-1em}
\end{figure*} 
}

In this section\footnote{All experiments were conducted using PyTorch; the code will be released on github upon acceptance of the paper.},
%is publicly available\ at \url{https://github.com/paper-submissions/mms-select}.}, 
%we report on our experiments that evaluated our PMM and the MMS selection methods for achieving a higher accuracy score as well as faster convergence, respectively. 
we report on a series of experiments designed to evaluate PMM's ability to achieve a higher accuracy score. %described by their authors. 
%For the selection scheme, we used uniform sampling from the dataset. We used several common datasets and neural network based models taken from vision and {\it natural language processing} (NLP) domains.
The experiments were conducted on commonly used datasets and DNN architectures, in {\it vision} and NLP realms.
For image classification, we used CIFAR10, CIFAR100 \cite{krizhevsky2009learning} and ImageNet \cite{imagenet_cvpr09} datasets.  For natural language inference, we used Question NLI (QNLI) \cite{wang2018glue}, MultiNLI (MNLI) \cite{williams2017broad} and Recognizing Textual Entailment (RTE) \cite{bentivogli2009fifth}. Lastly, for text classification and sentence similarity, we used the MSR Paraphrase Corpus (MRPC) \cite{dolan2005automatically},  Quora Question Pairs (QQP) \cite{chen2018quora}, and the Stanford Sentiment Treebank-2 (SST-2) \cite{socher2013recursive}.

\subsection{Image Classification}

For small-scale image classification, we used CIFAR10 and CIFAR100 datasets. These datasets comprise $32 \times 32$ color images from 10 or 100 classes, consisting of 50k training examples and 10k test examples. The last 5k images of the training set are used as a validation set, as suggested in common practice. For our experiments, we used ResNet-44 \cite{he2016deep} and WRN-28-10 \cite{zagoruyko2016wide} architectures. We applied the original hyperparameters and training regime using a batch size of 64. In addition, we used the original augmentation policy as described in \cite{he2016deep} for ResNet-44, while adding cutout \cite{devries2017improved} and auto-augment \cite{cubuk2018autoaugment} for WRN-28-10. Optimization was performed for 200 epochs (equivalent to 156k iterations) after which baseline accuracy was obtained with no apparent improvement. 

PMM was added to the objective function as an additional regularization term, where $\alpha$ is a trade-off factor between the cross-entropy loss and the regularization~\footnote{This formulation is equivalent to Equation~\ref{eq:d4}, where  $\alpha=\frac{1}{\lambda}$. It is preferred because it leads to multiplying the regularization term by a small number and keeping the scaling factor of ${\cal C}_i$ to be 1, thus avoiding gradient enlargement.}:
$$
\mathcal{L}(\theta) =  \alpha \sum_i {\cal R}_i + \sum_i {\cal C}_i
$$
%\newline
To find the optimal $\alpha$, we used a grid search and found that a linear scaling of $\alpha$ in the range of $[1\text{e-}5..1\text{e-}3]$ works best for  CIFAR10/100 and static $\alpha=1\text{e-}5$ works best for ImageNet. 

\begin{figure*}[tbh]
% \vspace{-1em}
    \centering
         \includegraphics[width=2\columnwidth]{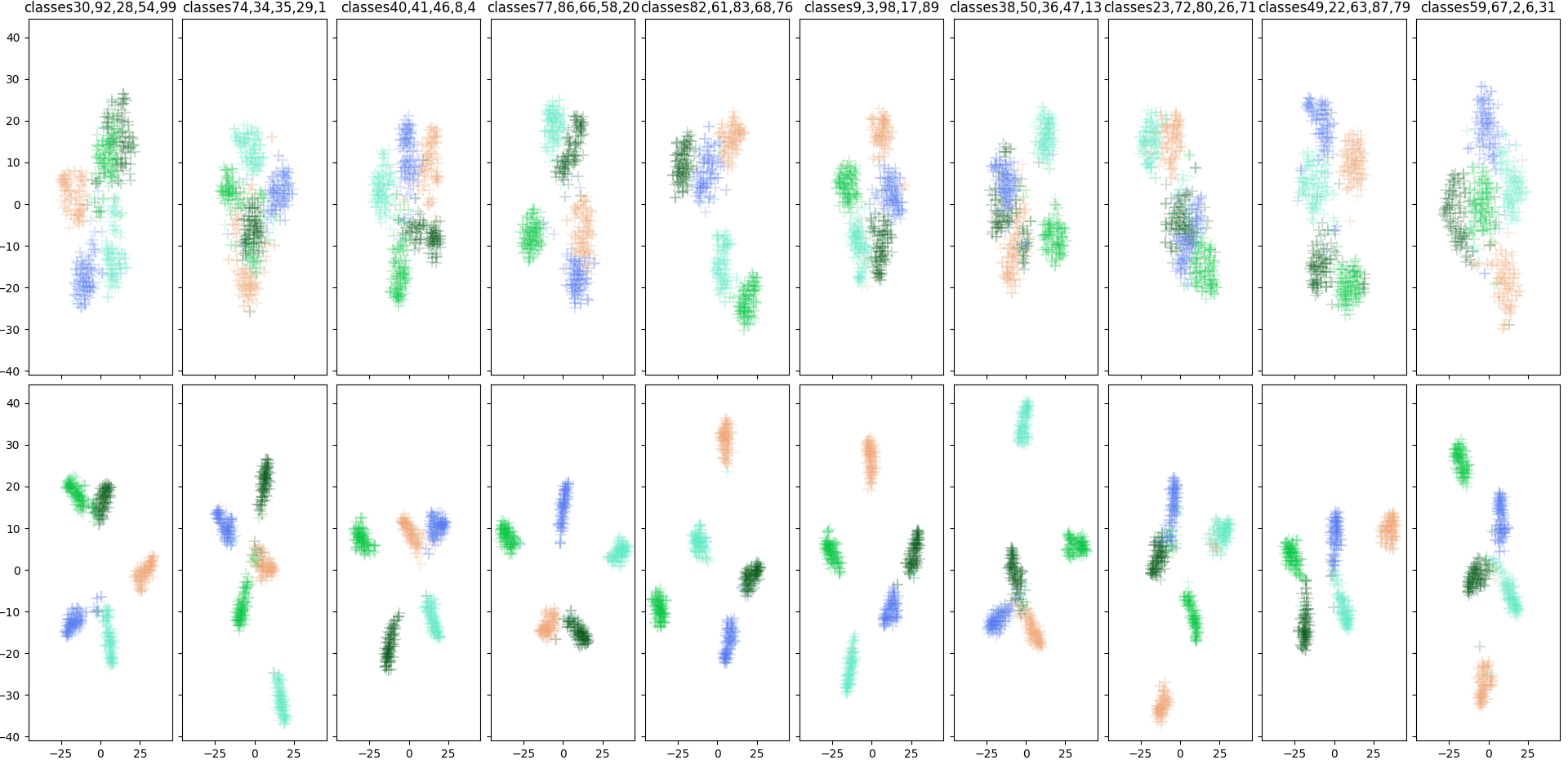}  
    % \qqqqqquad %add desired spacing between images, e. g. ~, \quad, \qquad, \hfill etc. 
     
     \caption{A scatter plot of the feature points in the last layer in WRN-28  trained on CIFAR-100.  Embedding into 2D was performed using t-SNE. Upper panel: Feature points using the baseline regularization scheme. Each plot indicates five classes randomly selected from 100 classes. Lower panel: Feature points of the same five classes using the PMM regularization scheme. }
    \label{TSNE_cifar100}
    % \vspace{-1em}
\end{figure*}

Figures~\ref{TSNE_cifar10_new} and \ref{TSNE_cifar100} show qualitative comparisons between the standard $L_2$  regularization and the suggested PMM scheme. They both
present scatter plots of feature points taken from the penultimate layer of the network (before the classification layer). These feature points were projected into 2D using t-SNE \cite{van2008visualizing}.  Figure~\ref{TSNE_cifar100} consists of a set of 10 plots for the standard regularization scheme (upper panel) and for the PMM regularization (lower panel). Each plot presents five randomly selected classes. In both figures (Figs~\ref{TSNE_cifar10_new} and \ref{TSNE_cifar100}), it is demonstrated that the feature points in the PMM scheme are clustered into distinct classes with large margins, while the baseline scheme presents tight and overlapping clusters with small margins. 

\begin{figure}[tbh]
% \vspace{-1em}
    \centering
         \includegraphics[width=0.9\columnwidth]{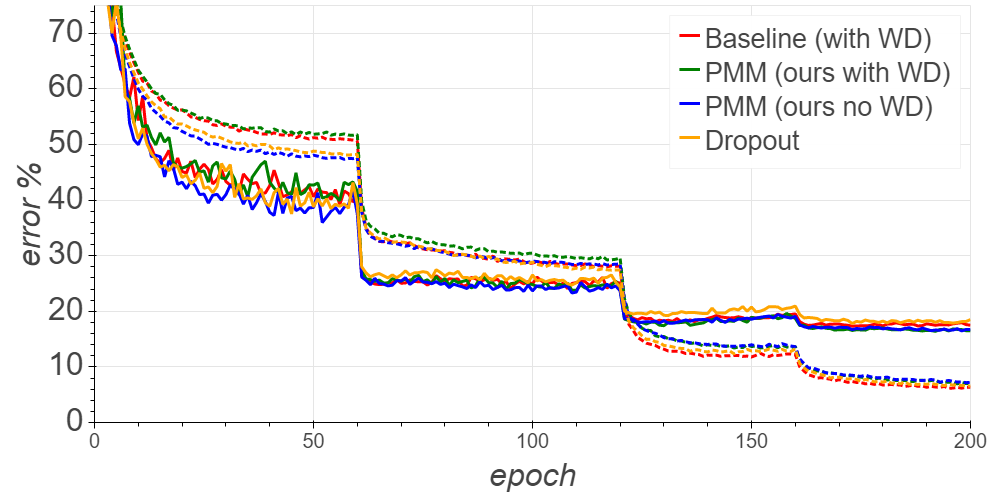}  
    % \qqqqqquad %add desired spacing between images, e. g. ~, \quad, \qquad, \hfill etc. 
        \includegraphics[width=0.9\columnwidth]{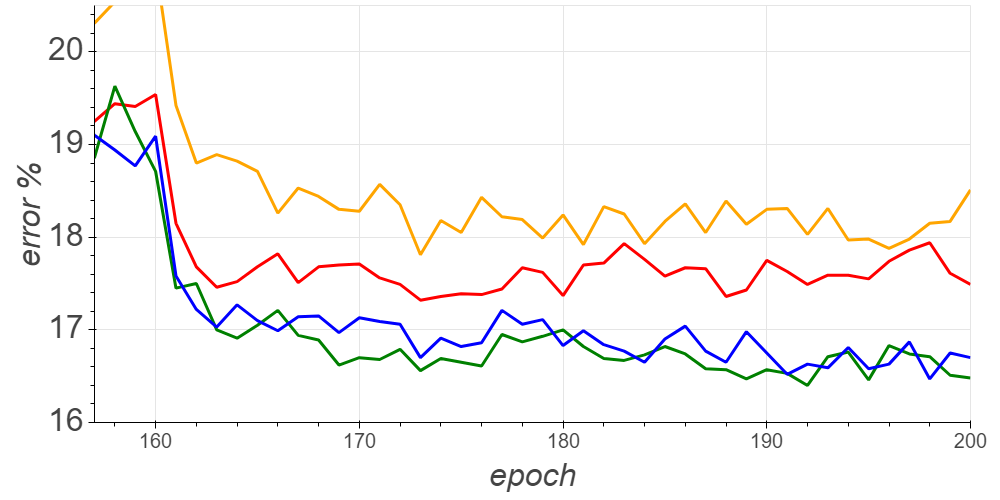} 
        % \caption{CIFAR100 on WRN28-10 val error with early drop regime}
     \caption{Training (dashed) and validation errors of CIFAR100 using the WRN28-10 neural network and comparing baseline training and our PMM approach. We use linear scale $\alpha$, starting with $1\text{e-}5$ up to $1\text{e-}3$.}
     \label{fig:lb}
    \label{compare_cifar100_mmp_ref}
    % \vspace{-1em}
\end{figure}

With respect to quantitative evaluation, Table~\ref{table:val_accuracy} demonstrates our final classification results on CIFAR-10 and CIFAR-100 when implementing PMM on several commonly used architectures. On CIFAR-10, we managed to improve baseline accuracy in ResNet-44 from $93.22\%$ to $93.83\%$ and in VGG, from $93.19\%$ to $93.34\%$. On CIFAR-100, we show a substantial increase using the WRN-28-20 model, raising its absolute accuracy by more than $1\%$.
\ignore{
Altogether, we observed a $2.67\%$ average decrease in error rates on all datasets.
}
In  Figure~\ref{compare_cifar100_mmp_ref} we compare the error rates of the PMM regularization scheme with weight decay and dropout regularization, which are commonly used in DNNs. From Figure~\ref{compare_cifar100_mmp_ref}, it is clear that the other regularization techniques do not match the accuracy gain of the PMM scheme. Additionally, adding weight decay to PMM does not improve the error rate.

For large-scale evaluation, we used the ImageNet dataset \cite{imagenet_cvpr09}, containing more than 1.2M images in 1k classes. We used MobileNet \cite{howard2017mobilenets} architecture and followed the training regime established by \cite{goyal2017accurate} (an initial learning rate (LR) of 0.1 is decreased by a factor of 10 in epochs 30, 60, and 80, for a total of 90 epochs). We used a batch size of $256$ and $L_2$ regularization over weights of convolutional layers as well as the standard data augmentation. Comparing the PMM scheme with the baseline scheme shows that accuracy increased from $71.17\%$ to $71.44\%$ (see Table~\ref{table:val_accuracy}).

\begin{table}[h]
\caption{Comparing accuracy results with PMM. The accuracy was measured using the top-1 criterion for CIFAR10/100 datasets.}
\begin{center}
\begin{tabularx}{0.5\textwidth}{@{}l*{3}{C}c@{}}
\toprule

Model & Dataset &  Baseline  & PMM \\
\midrule
ResNet-44 \cite{he2016deep} & CIFAR10 & 93.22\% & {\bf 93.83}\%   \\

VGG \cite{simonyan2014very} & CIFAR10 & 93.19\% & {\bf 93.34}\%  \\

WRN-28-10+auto-augment+cutout \cite{zagoruyko2016wide} & CIFAR100 & 82.51\% & {\bf 83.52}\%  \\
VGG+auto-augment+cutout & CIFAR100 & 73.93\% & {\bf 74.19}\%  \\

MobileNet \cite{howard2017mobilenets} & ImageNet & 71.17\% & {\bf 71.44}\% \\

\hline
\end{tabularx}
\label{table:val_accuracy}
\end{center}
\end{table}

%\subsection{Improving accuracy via PMM}

\ignore{
To examine our PMM scheme as described in detail in Section \ref{MMP_details}, we added it to the objective function as an additional regularization term. We used a trade-off $\alpha$ factor between the cross-entropy loss and the additional regularization as follows~\footnote{Equivalently to Equation~\ref{eq:d4} we used $\alpha=\frac{1}{\lambda}$ to multiply the regularization term by a small number and keep the scaling factor of ${\cal C}_i$ to be 1, to avoid gradient enlargement}:
}

%\
%\newline
%\newline

\subsection{Natural Language Classification Tasks}
To challenge our premise that we could achieve a higher accuracy score, we tested our PMM on an NLP-related model and datasets. In particular, we used the BERT\textsubscript{BASE} model \cite{devlin2018bert} with 12 transformer layers, a hidden dimensional size of 768 and 12 self-attention heads. Fine-tuning was performed using the Adam optimizer as in the pre-training, with a dropout probability of 0.1 on all layers. Additionally, we used an LR of $2\text{e-}5$ over three epochs in total for all the tasks. We used the original WordPiece embeddings \cite{wu2016google} with a 30k token vocabulary. For our method, similarly to the image classification task, we also used the $\alpha$ factor in the objective function, and found, via a grid search, $\alpha=1\text{e-}5$ to be the optimal value~\footnote{We applied $\alpha=1\text{e-}6$ only to evaluate our method's accuracy with the mismatched MNLI.}. 

We performed experiments on a variety of supervised tasks, specifically by applying a downstream task of fine-tuning natural language inference, semantic similarity, and text classification. All these tasks are available as part of the GLUE multitask benchmark \cite{wang2018glue}.

\paragraph{Natural Language Inference}
The task of natural language inference (NLI) or recognizing textual entailment means that when a pair of sentences are given, the classifier decides whether or not they contradict each other. Although there has been a lot of progress, the task remains challenging due to the presence of a wide variety of phenomena such as lexical entailment, coreference, and lexical and syntactic ambiguity. We evaluate our scheme on three NLI datasets taken from different sources, including transcribed speech, popular fiction, and government reports (MNLI), Wikipedia articles (QNLI) and news articles (RTE).

\ignore{
As shown in Table~\ref{table:val_accuracy}, our scheme using the regularization term outperformed baseline results on all the three tasks. We achieved absolute improvement of up to $1.44\%$ on RTE and a relative change in error of $4.53\%$. On QNLI and MNLI we also achieved higher scores of $91.48\%$ (accuracy) and $84.70\%$ (F1), outperforming the baseline results by $0.42\%$ and $0.2\%$, respectively.
}
As shown in Table~\ref{table:val_accuracy_nlp}, our PMM scheme outperformed baseline results on all three tasks. Specifically, on RTE we obtained an absolute improvement of nearly $1.5\%$ (from $68.23\%$ accuracy to $69.67\%$) with respect to the baseline score.

\paragraph{Semantic Similarity}
This task involves predicting whether two sentences are semantically equivalent by identifying similar concepts in both sentences. 
It can be challenging for a language model to recognize syntactic and morphological ambiguity as well as compare the same ideas using different expressions or the other way around. 
We evaluated our approach on QQP and MRPC downstream tasks, outperforming baseline results as can be seen in Table \ref{table:val_accuracy_nlp}. On MRPC in particular, we achieved a $0.75\%$ improvement over the baseline, which is a relative change of more than $8\%$. 

\paragraph{Text Classification}
Lastly, we evaluated our method on the Stanford Sentiment Treebank (SST-2), 
which is a binary single-sentence classification task consisting of sentences
extracted from movie reviews with human annotations regarding their sentiment. 
Here too, our approach outperformed the baseline by a  small increase in the accuracy.

Overall, applying PMM boosted the accuracy in all the reported tasks, indicating that our approach works well for different tasks from various domains.

\begin{table}[h]
\caption{Comparing accuracy results - PMM vs. the baseline. F1 scores are reported for QQP and MRPC. For MNLI, we report the average of the matched (with $\alpha=1\text{e-}5$) and mismatched subsets (with $\alpha=1\text{e-}6$) for both, the baseline and our PMM.}
\begin{center}
\begin{tabularx}{0.5\textwidth}{@{}l*{3}{C}c@{}}
\toprule

Model & Dataset &  Baseline  & PMM \\
\midrule

  & MNLI & 84.5\% & {\bf 84.70}\%    \\
  & QNLI & 91.06\% & {\bf 91.48}\%   \\
  BERT\textsubscript{BASE} \cite{devlin2018bert}  & RTE & 68.23\% & {\bf 69.67}\%   \\

 & QQP & 87.9\% & {\bf 88.04}\%   \\
    
     & MRPC & 90.68\% & {\bf 91.43}\%  \\
     
  & SST-2 & 92.08\% & {\bf 92.43}\%    \\

\hline
% \multicolumn{4}{l}{$^{\mathrm{a}}$Sample of a Table footnote.}
\end{tabularx}
\label{table:val_accuracy_nlp}
\end{center}
\end{table}

\section{Discussion}
\label{sec:Discussion}

We studied a multi-class margin analysis for DNNs and used it to devise a novel regularization term we call {\it Pairwise Margin Maximization} (PMM). %Similarly to previous formulations, the 
The PMM term aims at increasing the margin induced by the classifiers, and it is derived directly, for each sample, from the true class and its most competitive class. 
Note that the standard weight decay or $L_2$ norm regularization scheme aims at maximizing the margins along the one-vs-all decision boundaries. 
In contrast, the PMM strives to maximize the margin along the one-vs-one decision boundaries, and we argue that this is the preferred multi-class scheme. 
\ignore{
We claim that the standard weight decay, or $L_2$ norm regularization scheme, aims at maximizing the margins along the one-vs-all decision boundaries which are not the correct boundaries to observe
due to the fact that a class decision is made by comparing the maximal score against the other class scores. 
Therefore, the one-vs-one decision boundaries are the ones that the maximal margins should be applied to.
}

Another difference between PMM and common regularization terms is that PMM is scaled by $\| \phi_{max}\|$, 
which is the maximal norm of the samples in the feature space.
This ensures a meaningful increase in the margin that is not induced by a simple scaling of the feature space.  Lastly, since the PMM term is added to each sample, PMM is formulated and performed over the margin distribution to compensate for class imbalance in the regularization term. PMM can be incorporated with any %empirical risk 
loss, i.e., is not restricted to hinge loss or cross-entropy losses. 
%And indeed, using PMM, we demonstrate improved accuracy over a set of experiments in images and text. 
Using PMM, we were able to demonstrate improved accuracy over a set of experiments in images and text. 

Similarly to \cite{jiang2019predicting}, PMM can be implemented at other layers in the deep architecture. This enables maximal margins that directly impact training at all levels. The additional computation associated with such a framework makes it less appealing from an efficiency perspective, which may be compensated %for 
by the gain in accuracy. The design of such additional PMM terms is left for further study.

\bibliography{main}

\bibliographystyle{IEEEtran}

\end{document}